\documentclass[letterpaper, 10 pt, conference]{src/ieeeconf}  

\IEEEoverridecommandlockouts                              

\usepackage{graphics} 
\usepackage{epsfig} 
\usepackage{times} 
\usepackage{amsmath} 
\usepackage{amssymb}  
\usepackage{colortbl}
\usepackage[hidelinks]{hyperref}
\usepackage{usebib}
\usepackage[capitalise]{cleveref}
\usepackage{colortbl}
\usepackage{siunitx}
\DeclareSIUnit\mph{mph}
\usepackage{dblfloatfix}
\usepackage{soul}

\usepackage{makecell}
\usepackage{booktabs}

\usepackage{todonotes}

\title{\LARGE \bf
Experience Filter:
Using Past Experiences on \\ Unseen Tasks or Environments}

\author {
    Anil Yildiz\textsuperscript{\rm 1},
    Esen Yel\textsuperscript{\rm 1},
    Anthony L. Corso\textsuperscript{\rm 1},
    Kyle H. Wray\textsuperscript{\rm 1, 2},
    Stefan J. Witwicki\textsuperscript{\rm 2}, 
    and Mykel J. Kochenderfer\textsuperscript{\rm 1}
\thanks{\textsuperscript{\rm 1}
Stanford Intelligent Systems Laboratory, 496 Lomita Mall, Stanford, CA, USA. Emails: \texttt{\{yildiz, esenyel, acorso, kylewray, mykel\}@stanford.edu}}%
\thanks{\textsuperscript{\rm 2}
Alliance Innovation Laboratory Silicon Valley, Santa Clara, CA, USA.\newline Email: \texttt{stefan.witwicki@nissan-usa.com}}%
}

\usepackage{bbm}
\usepackage[capitalise]{cleveref}
\usepackage{graphics} 
\usepackage{epsfig} 
\usepackage{amsmath} 
\usepackage{subfigure}

\newcommand{\Space}[1]{\mathcal{#1}}
\newcommand{\T}{T(s' \mid s,a)}

\newcommand{\Z}{Z(o \mid s', a)}
\newcommand{\R}{R(s, a)}
\newcommand{\Real}{\mathbb{R}}
\newcommand{\Int}{\mathbb{Z}}
\newcommand{\Dir}{\text{Dir}}
\newcommand{\onevector}{\mathbf{1}}

\newtheorem{defn}{Definition}[section]

\DeclareMathOperator*{\argmin}{arg\,min}
\DeclareMathOperator*{\argmax}{arg\,max}
\newcommand{\minititle}[1]{\textbf{\textit{#1:}}}

\usepackage[backend=bibtex,style=ieee,mincitenames=1,maxcitenames=2,maxbibnames=20,url=false,doi=false]{biblatex}
\usepackage{flushend}

\begin{document}

\maketitle
\thispagestyle{empty}
\pagestyle{empty}

\begin{abstract}

One of the bottlenecks of training autonomous vehicle (AV) agents is the variability of training environments.
Since learning optimal policies for unseen environments is often very costly and requires substantial data collection,
it becomes computationally intractable to train the agent on every possible environment or task the AV may encounter.

This paper introduces a zero-shot filtering approach to interpolate learned policies of past experiences to generalize to unseen ones.
We use an experience kernel to correlate environments.
These correlations are then exploited to produce policies for new tasks or environments from learned policies.
We demonstrate our methods on an autonomous vehicle driving through T-intersections with different characteristics, where its behavior is modeled as a partially observable Markov decision process (POMDP).
We first construct compact representations of learned policies for POMDPs with unknown transition functions given a dataset of sequential actions and observations.
Then, we filter parameterized policies of previously visited environments to generate policies to new, unseen environments.
We demonstrate our approaches on both an actual AV and a high-fidelity simulator.
Results indicate that our experience filter offers a fast, low-effort, and near-optimal solution to create policies for tasks or environments never seen before.
Furthermore, the generated new policies outperform the policy learned using the entire data collected from past environments, suggesting that the correlation among different environments can be exploited and irrelevant ones can be filtered out.

\end{abstract}


\section{Introduction}
Designing efficient and safe planning strategies for autonomous vehicles is generally challenging due to uncertainty. One of the principled ways of modeling the planning problem under uncertainty is using POMDPs~\cite{kaelbling1998planning}, which have been successfully applied in autonomous driving~\cite{cailets, sunberg_pomdp}.  The POMDP formulation includes a state transition model which represents the environment dynamics. When this model is not readily available for a given problem, machine learning techniques can be used to learn the environment models from data collected at design time. Although reinforcement learning algorithms have been shown to produce effective policies for environments that they are trained upon~\cite{kochenderfer2015_dmu}, they typically need to be retrained to be deployed in new environments.
To overcome this limitation, models or policies previously learned need to be transferred to new, unforeseen tasks.

\begin{figure}[t!]
\centering
\includegraphics[width=\columnwidth, trim={0 0 0 0cm},clip]{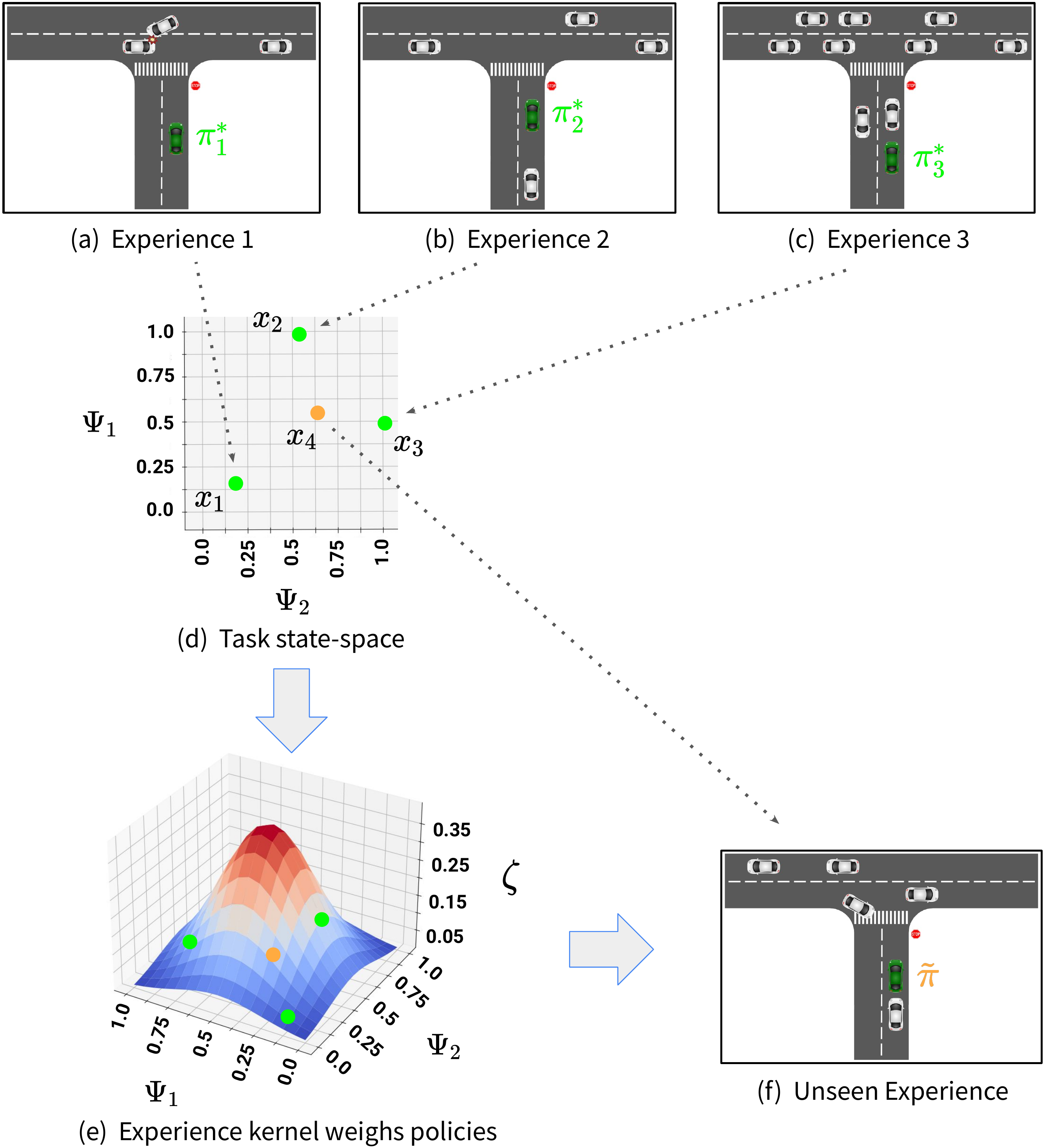}
\caption{(a)-(d) Different tasks or environments $x_i\in \Space{X}$ (and their policies $\pi_i^*$) can be compactly represented through parameterization.
(e) Here, we define the relationship between a new environment (parameterized by $x_4$) and previously visited environments (parameterized by $x_1,x_2,x_3$) using a kernel $\zeta(x_m, x_4)$, $m=\{1,2,3\}.$ \linebreak
(f) An \textit{experience filter} exploits this relation to generate a policy for $x_4$ given policies trained on different tasks or environments.}
    \label{fig:exp-filter-intro}
    \vspace{-5pt}
\end{figure}

In this work, we introduce the concept of an experience filter, illustrated in \cref{fig:exp-filter-intro}, to reason about new tasks or environments using past experiences. 
The goal of the experience filter is to filter optimal/learned policies of previously visited environments to generate policies to ones never seen before.
By doing so, we eliminate the need to train an autonomous vehicle's policy on a vast range of different tasks, but rather, interpolate existing policies to unseen similar tasks, eliminating the need to collect any new data.
Our experience filter (EF) approach allows fast, low-effort, and near-optimal solutions to create
policies without requiring to train for tasks or environments not seen before.
Furthermore, our results also indicate that using the EF approach to correlate and filter previous experiences through a kernel, as depicted in \cref{fig:exp-filter-intro}, yields higher performance than simply using the entire data collected in an attempt to train an all-for-one policy.

We assume that both policies and environments can be represented by low-dimensional, easily accessible parameter vectors.
For example, the behavior of an AV navigating along a highway can be influenced by factors such as traffic density, number of lanes, upcoming intersections, etc. 
Our proposed experience filter takes environment representations and parameterized policies as an input, and using an experience kernel, outputs a new policy for an unseen environment through Bayesian reasoning between the learned and unseen environment parameters. 

The contributions of this paper are twofold: (1) we introduce a new technique to learn and parameterize policies of a POMDP with an unknown transition function from existing data, and
(2) we propose an experience filter to efficiently plan in unseen environments by interpolating past knowledge. 
We demonstrate our approach on a problem of autonomously driving through T-intersections, and we include demonstrations on both an actual AV and a high-fidelity simulator CARLA~\cite{carla-sim}.
For simulation results, we benchmark our experience filter approach
using the following performance metrics: collision risk, discomfort during driving, and task completion time.

The paper is organized as follows: \Cref{sec:rel_work} gives an overview of transfer learning techniques. \Cref{sec:background} gives a brief background on Dirichlet distributions, POMDPs, and MODIA and \cref{sec:problem} defines the problem. \Cref{sec:method} presents our learning approach and experience filter.  \Cref{sec:results} summarizes results. Finally, \cref{sec:conclusion} draws conclusions and discusses future work.

\section{Related Work} \label{sec:rel_work}
The problem of how to reduce training effort for new tasks by leveraging the knowledge from existing tasks has been addressed by transfer learning techniques. 
Existing work on transfer learning techniques largely focuses on transferring knowledge from previously seen \emph{source} tasks to new tasks~\cite{zhuang2020comprehensive} and 
can be divided into the following three categories.

\paragraph{Model-free transfer} 

Transfer learning for model-free reinforcement learning algorithms involves transferring knowledge between tasks in the form of experience samples, policies, or value functions.
    Progressive networks~\cite{rusu2016progressive} is an approach to transfer learning where additional neurons are added to a network for each new task that is learned. Previously learned network weights are frozen, and the new neurons are connected so as not to change the output of the network on previous tasks. 
The attend, adapt, and transfer approach~\cite{rajendran2017attend} uses a set of attention weights to combine the output of a set of source policies to achieve good performance on a new task. To avoid negative transfer, a new policy is learned from scratch and combined with the source policies through an additional weight. The attention weights can be learned from a small amount of data, allowing for fast adaptation to new environments. Transfer learning with model-free deep reinforcement learning is also applied to task transfer for autonomous driving decision making problems~\cite{shu-itvt-2022}.

\paragraph{Model-based transfer}

Prior work on transfer in a model-based setting assumes there are unobserved parameters that describe each task and seek to learn a model of these parameters from data~\cite{chrisman1992reinforcement, choi1999environment}.
    Recent work~\cite{killian2017robust} assumes a low-dimensional latent task identifier which is inferred online using a Bayesian neural network, allowing for fast adaptation to new problems.
Hidden parameter MDPs were extended to allow for variations in the reward function~\cite{perez2020generalized}.
    Storing experience samples from previous source tasks and constructing a model for the new task with an inter-task mapping has also been done~\cite{taylor2008transferring}.
Our approach is similar to the aforementioned work in that transfer to a new task is done by combining previously learned models. Both model-based and model-free approaches, however, require training on the new task, whereas we consider the case of zero-shot transfer in this paper.

\paragraph{Zero-shot transfer}

Zero-shot transfer~\cite{wang2019survey} is often preferred when there is a known relationship between tasks. For example, when the source task is a simulated version of the real target task, unsupervised pre-training has shown to be effective for enabling zero-shot transfer~\cite{higgins2017darla}. Alternatively, automated measures of MDP similarity can be used~\cite{ammar2014automated}. Zero-shot policy transfer along with the robust tracking controllers to tackle the source to target modeling gap is applied in robotics~\cite{harrison2020adapt} and autonomous driving problems~\cite{xu2018zero}.
In our setting, we exploit a parametric connection between previously visited environments, using this relation to hypothesize policies for unseen environments.

\section{Background} \label{sec:background}

\subsection{Dirichlet Distributions}
Dirichlet distribution $ \Dir\left(\boldsymbol\alpha_{1:n} \right) $ is parameterized by $ \boldsymbol\alpha_{1:n} \in \Real_{\geq 0}^n $ which can be treated as pseudo-counts of different outcomes. The density of a Dirichlet distribution is given by
\vspace{-5pt}
\begin{align*}
    P
    \left(
    \boldsymbol\theta_{1:n} \mid \boldsymbol\alpha_{1:n}
    \right)
    =
    \dfrac{\Gamma\left( \sum_{i=1}^n\alpha_i \right)}{\prod_{i=1}^n \Gamma\left( \alpha_i \right)} \prod_{i=1}^n\theta_i^{\alpha_i -1}
\end{align*}
where $\Gamma$ is the gamma function~\cite{kotz2004continuous}.

If the prior is a Dirichlet distribution and the event $ i $ is observed $m_i \in \boldsymbol m_{1:n}$ times, then the posterior is also a Dirichlet distribution~\cite{kochenderfer2015_dmu}:
\begin{align*}
    \boldsymbol\theta_{1:n} \mid \boldsymbol\alpha_{1:n}, \boldsymbol m_{1:n}
    \sim
    \Dir
    \left(
    \boldsymbol\alpha_{1:n} + \boldsymbol m_{1:n}
    \right).
\end{align*}



\subsection{Partially Observable Markov Decision Processes}
A sequential decision making problem can be modeled as a partially observable Markov decision process. A POMDP model is represented by a tuple $ \langle \Space{S}, \Space{A}, \Space{O}, T, Z, R, \gamma \rangle $ where $ \Space{S} $ is a finite set of states, $ \Space{A} $ is a finite set of actions, and $ \Space{O} $ is a finite state of observations. The system takes an action $ a \in \Space{A} $ from state $ s \in \Space{S} $ and transitions to the next state $ s' \in \Space{S} $ according to the probabilistic transition function $ T(s' \mid s,a) = P(s'\mid s,a) $ which models the environment dynamics. From state $ s' $ the system obtains observation $ o \in \Space{O} $ according to the observation function $ Z(o\mid s', a) = P(o\mid s',a) $ and receives a reward according to the reward function $ R(s,a) $. As the system does not have access to the true world states, it maintains a belief $ b(s) $ over possible states.
A discount factor $\gamma \in [0,1]$ may also be used to prioritize earning rewards sooner than later. 


The goal of the system is to find a policy $ \pi^* $ to maximize the expected total discounted reward starting from its belief $ b $, which can be exactly calculated using the relation
\begin{align}
\pi^*(b)
=\argmax_a
\sum_{s}
b(s)Q(s,a)
\label{eqn:pi_star}
\end{align}
where
\vspace{-5pt}
\begin{align}
V^*(b) = \max_a
\sum_{s} b(s) Q(s,a)
\label{eqn:V_ba}
\end{align}
\vspace{-5pt}
\begin{align}    
\begin{split}
    Q(s & , a) = R(s,a) \\ & + \gamma \sum_{s'} T(s'\mid s,a) \sum_o Z(o\mid s',a) V^*(b'_{(b,a,o)})
\end{split} 
\label{eqn:Q_sa}
\end{align}
and $b'_{(b,a,o)}$ is the updated belief.

\setcounter{equation}{4}
\begin{table*}[!t]
\normalsize
\vspace{-10px}
\centering
\begin{equation}\label{eq:m_sas}
  \begin{gathered}
    \boldsymbol m_{(s,a)}
    =
    \begin{bmatrix}
    m_{(s,a,s_1')} \\    m_{(s,a,s_2')} \\
    \vdots \\ 
    m_{(s,a,s_{\vert \Space{S} \vert}')}
    \end{bmatrix}
    \quad \text{where} \quad
    m_{(s,a,s')}
    =
    \sum_{S_k \in \Space{D}_i}
    \sum_{C_j \in S_k}  
    \sum_{{(s_{ijk}, a_{ijk}, s'_{ijk})} \in C_j}
    \int_{t=0}^{t=\tau_k} \onevector
    \begin{Bmatrix}
    s_{ijk}(t) = s\\ a_{ijk}(t) = a\\ s'_{ijk}(t) = s'
    \end{Bmatrix}
    dt
    \end{gathered}
\end{equation}
\vspace{-10px}
\end{table*}

For fully observable states (MDP), the 
belief and observation terms in \cref{eqn:pi_star,eqn:V_ba,eqn:Q_sa} drop out, and the optimal solution can be computed tractably using Bellman updates~\cite{bellman2015applied}. 
However, finding a solution to a POMDP using dynamic programming is computationally very expensive, and thus, approximate solutions are often used instead~\cite{kochenderfer2015_dmu}. In this work, we use the QMDP~\cite{hauskrecht2000value} offline solver, and assume the resulting policy to be near-optimal.




\subsection{MODIA} \label{subsec: MODIA}

The multiple online decision-components with interacting actions (MODIA) framework \cite{ijcai2017-664} aims to solve complicated real world decision making problems in a scalable way by separating them into subproblems.
Instead of constructing a single POMDP that accounts for all of the other agents in a domain, MODIA instantiates multiple smaller POMDPs for each agent interaction.
This formulation inherently assumes that most agents act independently from one another.
At each timestep, the safest action among all POMDPs is selected for execution.
MODIA is utilized during the experimentation.

\section{Problem Definitions}  \label{sec:problem}




In this study, our goal is to efficiently generate new  policies for unseen driving tasks or environments using past experiences. To achieve this, we need to represent policies by a compact representation, and then find a way to transfer the policies. These two problems are defined as follows:

\textbf{\textit{Problem 1: Parameterizing Policies:}} Given a driving dataset $\Space{D}_i $ containing observation triplets $ (o^{t}_i, a^{t}_i, o^{t+1}_i)$ obtained from $N$ different driving environments visited at times $t \in [0, \tau_i]$, and a policy $\pi_i^*$ learned for environment $i$, what is the compact representation $par(\pi_i^*)$ of this policy? 

\textbf{\textit{Problem 2: Experience Transfer:}} Given a set of policy representations $ par(\pi^*_i), \ i = 1, \dots, N  $ learned for $N$ visited environments, how can we compute a reasonable policy for an environment that was never seen before?

In this paper, we investigate the behavior of an AV at T-intersections.
We model the AV's behavior as a POMDP, and assume that the $Z(o\mid s', a)$ and $R(s,a)$ functions are known.
In the first portion of this paper, we focus on learning the transition function $T(s' \mid s,a)$ for T-intersections of different characteristics. 
Given that $T$, $Z$ and $R$ are learned or known, the optimal policy of the POMDP can therefore be solved for from \cref{eqn:pi_star,eqn:V_ba,eqn:Q_sa}.
In the second portion of this paper, we introduce a framework that allows us to deduce a policy for a T-intersection type that was never seen before, by generalizing a handful number of previously learned policies.
We discuss our approaches to both of these problems in \cref{sec:method}.



\section{Efficient Transition Function Learning and Transferring} \label{sec:method}

In this section, we first discuss how the policy is learned for an individual T-intersection.
Then, the experience filter, that allows computing a policy for unseen T-intersection types, is formulated.

\subsection{Learning and Parameterizing the Policy} \label{subsec:learn_pol}

The $\Space{S}$, $\Space{A}$, $\Space{O}$ of the POMDP used to model the AV's behavior at a T-intersection while considering a single rival vehicle is defined as follows.

\minititle{State Space $\Space{S}$} 
Each state $s \in \Space{S}$ is a 5-dimensional
$(pos_{ego}, \ sgt_{ego}, \ pos_{rival}, \ blk_{rival}, \ aggr_{rival})$ vector.
Here, $pos_{ego},\ pos_{rival} \in \{\texttt{before}, \ \texttt{at}, \ \texttt{inside}, \ \texttt{after}\}$ denote the position of the ego and rival vehicles with respect to the stop sign of the T-intersection, respectively,
$sgt_{ego} \in \{\texttt{yes}, \ \texttt{no}\}$ denotes whether or not the ego vehicle has a clear line of sight,
$blk_{rival} \in \{\texttt{yes}, \ \texttt{no}\}$ denotes whether or not the rival vehicle is blocking the ego vehicle's path,
$aggr_{rival} \in \{\texttt{cautious}, \ \texttt{normal}, \ \texttt{aggressive}\}$ denotes the aggressiveness level of the rival vehicle.
Since all five dimensions of the state space are discretized, there are a total of $192$ possible states in $\Space{S}$.

\minititle{Action Space $\Space{A}$} 
Consists of three permissible actions of the AV, which are  $\{\texttt{stop}, \ \texttt{edge}, \ \texttt{go}\}$.

\minititle{Observation Space $\Space{O}$} 
The observation space is equivalent to the state space $\Space{S}$, and therefore also has a cardinality of $192$. Note that observation $o$ is noisy over the true state $s$.

In this study, we assume that the observation and reward functions, $\Z$ and $\R$, are known and constant across different types of T-intersections.
This is a reasonable assumption since $Z$ inherently captures the sensing accuracy, and $R$ describes driving preferences, both of which can be quantitatively modeled a priori.
Therefore, by inspecting \cref{eqn:pi_star,eqn:V_ba,eqn:Q_sa}, learning a policy for a specific T-intersection reduces to an accurate representation of the transition function $\T$.


In this study, we represent the learned transition function $\tilde T$ as the posterior Dirichlet distribution:
\begin{align*}
    \boldsymbol \theta(s,a)
    \sim
    \Dir\left(
    \boldsymbol\alpha_{(s,a)} + \boldsymbol m_{(s,a)}
    \right) \tag{4}
\end{align*}
where
$\boldsymbol\theta(s,a) \in \left[0,1\right]^{\vert \Space{S} \vert}$ is a vector whose each element is
$\theta(s,a) = \tilde T(s' \mid s,a), \ \forall s'\in \Space{S}$, given $s,a$.
Parameters $\boldsymbol\alpha_{(s,a)} \in \Real_{\geq 0}^{\vert \Space{S} \vert}$, and
$\boldsymbol m_{(s,a)} \in \Int_+^{\vert \Space{S} \vert}$ describe the prior and likelihood pseudo-counts, respectively.

A dataset $\Space{D}_i$ is collected as a specific intersection $i$ is driven through $d$ times.
We call each drive through an intersection a \textit{scenario}, hence, there are $d$ scenarios in $\Space{D}_i$.
During each scenario $S_k$, the ego vehicle interacts with multiple rival cars and receives observation triplets, each of them labeled $C_j$.
As a part of the MODIA definition (\cref{subsec: MODIA}), a separate PODMP is instantiated for each rival car at the T-intersection.
These instantiated POMDPs may use a baseline or expert policy (if available). 
As a result, separate $(s,a,s')$ triplets are recorded for each rival vehicle.
Therefore, the likelihood pseudo-counts $\boldsymbol m_{(s,a)}$ can be formulated as
in \cref{eq:m_sas}
where 
$\tau_k$ denotes the total time taken in scenario $S_k$, and
$\onevector\{\cdot\}$ outputs a scalar $1$ if all expressions inside the curly braces are true, and $0$ otherwise.
However, due to partial observability, we cannot directly observe the states, but rather, receive observations from the world, making \cref{eq:m_sas} inaccessible for POMDPs.
Furthermore, the integral in \cref{eq:m_sas} is often intractable.
As a remedy, we use the \textit{most likely state} with respect to the observations received, and discretize the scenarios into small timesteps:
\begin{align}
    m_{(s,a,s')}
    =
    \sum_{S_k \in \Space{D}_i}
    \sum_{C_j \in S_k}  
    \sum_{{(o_{ijk}^{t}, a_{ijk}^{t}, o_{ijk}^{t+1})} \in C_j}  
    \sum_{t=0}^{\tau_k} \onevector
    \begin{Bmatrix}
    \tilde s_{ijk}^{t} = s\\ a_{ijk}^{t} = a\\ \tilde s_{ijk}^{t+1} = s'
    \end{Bmatrix} \label{eq:m_sas2}
\end{align}
where
\begin{align*}
    \tilde s^{t+1}_{ijk} & = \argmax_{s} b_{ijk}^{t+1}(s) \\
    b_{ijk}^{t+1}(s) & = P(s \mid b_{ijk}^t,a_{ijk}^t,o_{ijk}^t).
\end{align*}

We represent a learned policy $\pi^*$ through a set of parameters $par(\pi^*)$. For POMDP representations, through \cref{eqn:pi_star,eqn:V_ba,eqn:Q_sa}, the optimal policy depends on $T$, $Z$ and $R$. 
As previously mentioned, we are treating $Z$ and $R$ to be known a priori, and assumed to be constant across different T-intersection characteristics.
Hence, we can readily accept the parameters that describe a learned policy in our AV domain to be $\T, \ \forall a \in \Space{A}, \ \forall s',s \in \Space{S}$.
From this logic, 
$par(\pi_i^*)$ for intersection $i$ is distributed by the product
\begin{align}
    par(\pi_i^*)
    \sim
    \prod_s \prod_a
    \Dir\left(
    \boldsymbol\alpha_{(s,a)} + \boldsymbol m_{(s,a)}
    \right)
\end{align}
and contains $\vert \Space{S} \vert^2 \times \vert \Space{A} \vert = 192^2 \times 3 = 110,592$ parameters.

\subsection{Experience Filter} \label{subsec:expfilter}

We first define the environment state-space $\Space{X}$ and experience kernel $\zeta(\cdot)$ as follows.

\defn \label{defn:X}
Let a set of parameters $(\psi_1, \cdots, \psi_K)$ where $\psi_k \in \Psi_k, \ k=1,...,K$ efficiently represent an environment.
Then, an environment state-space is the Cartesian product $\Space{X} = \Psi_1 \times \cdots \times \Psi_K$ which can be defined~as
\begin{align}
    \Space{X}
    =
    \left\{
    (\psi_1, \cdots, \psi_K) \mid 
    \psi_k \in \Psi_k, \ k=1,...,K
    \right\}.
\end{align}

\defn \label{defn:zeta}
Given an $K$-dimensional environment state-space $\Space{X}$, the experience kernel $\zeta(x_n, x_k) $ outputs the correlation between two environment states $x_n, x_k \in \Space{X}$ while satisfying
\begin{align}
    \sum_{k=1}^K \zeta(x_n, x_k) = 1 \quad \forall n \neq k.
\end{align}
Using Defs. \ref{defn:X} and \ref{defn:zeta}, we can define the experience filter.

\defn
Given the environment states visited and recorded $ \Space{D}_\Space{X} \subset \Space{X} $, and parameterized learned policies computed for these environments $\Space{D}_\Phi = \left[par(\pi^*_i)\right]_{i=1}^N$, an experience filter for an unseen environment state $x$ is formulated as
\begin{align} \label{eq:EF}
EF(x, \Space{D}_\Space{X}, \Space{D}_\Phi) & = \boldsymbol\zeta(x, \Space{D}_\Space{X})^T \Space{D}_\Phi\\
& = \sum_{i=1}^N \zeta(x, x_i) \ par(\pi_i^*). \notag
\end{align}
Here, $\boldsymbol\zeta(x, \Space{D}_\Space{X})^T$ is an $N$-dimensional vector whose elements are $\zeta(x, x_n), \ \forall x_n \in \Space{D}_\Space{X}$.
Intuitively, \cref{eq:EF} takes a weighted average of the parameters of learned policies to previously seen environments.
The resulting value is used as the parameters of the articulated policy for the unseen environment $x$, which for our AV domain, acts as the transition function for any POMDP instantiated in $x$.

\section{Experimentation} \label{sec:results}
In this section, we first demonstrate learning policies for visited environments using the approach described in \cref{subsec:learn_pol}.
Then, using the CARLA simulator~\cite{carla-sim}, we demonstrate how our experience filter from \cref{subsec:expfilter} can generalize previously learned policies to unseen environments states.

\begin{figure}[t]
    \centering
    \centering
    \subfigure[Real AV at intersection]{\includegraphics[width=0.48\columnwidth]{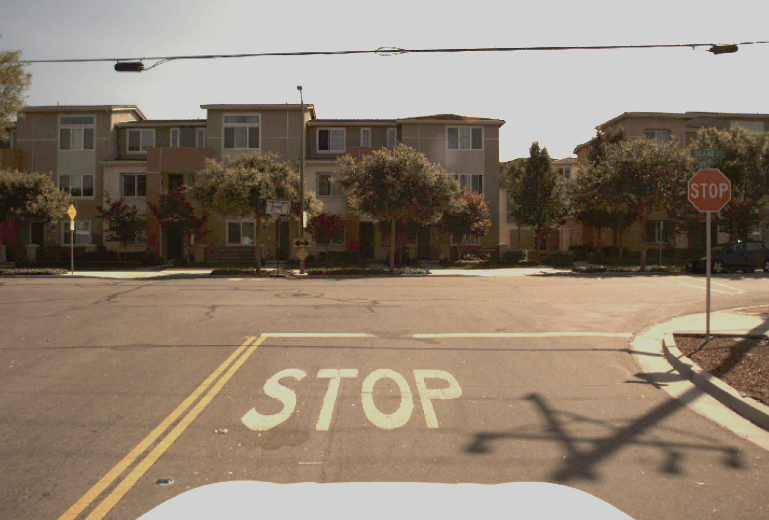}}
    \subfigure[Speed profile]{\includegraphics[width=0.48\columnwidth]{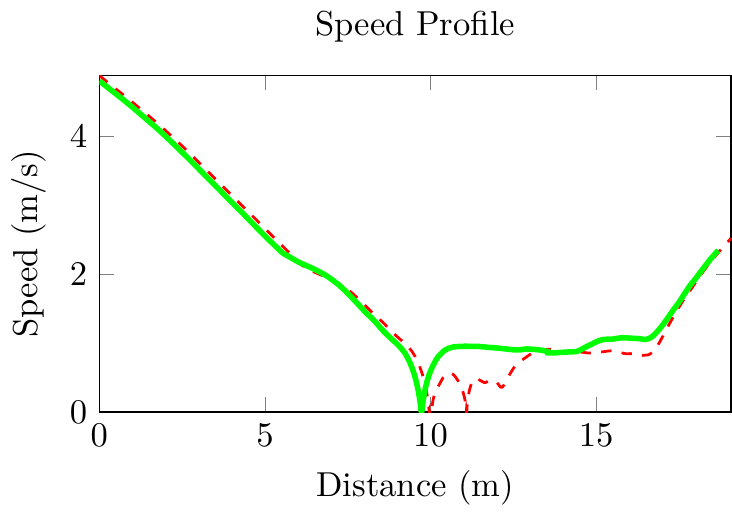}}
    \subfigure[Time profile]{\includegraphics[width=0.48\columnwidth]{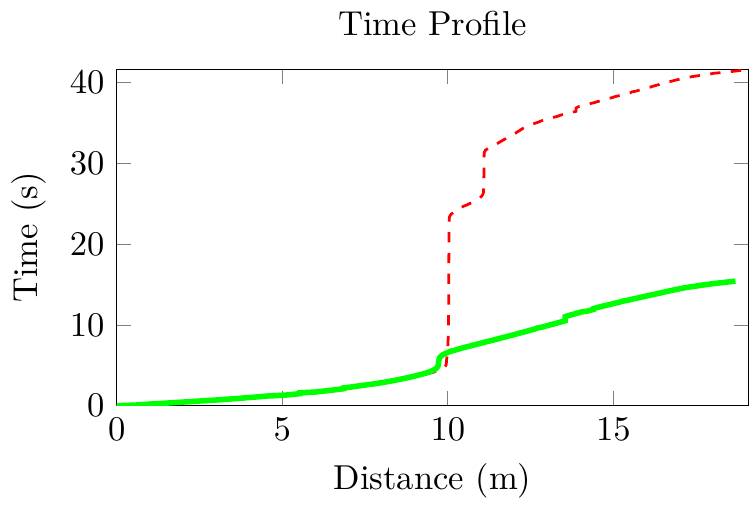}}
    \subfigure[Visibility profile]{\includegraphics[width=0.48\columnwidth]{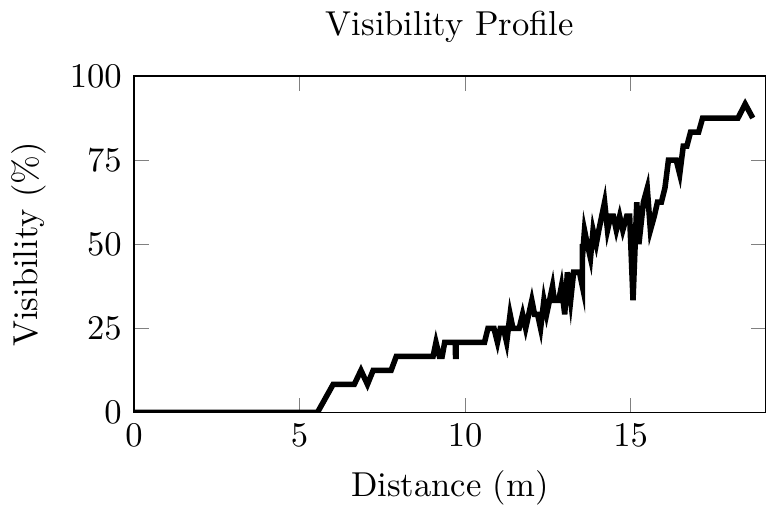}}
    \caption{A fully operational AV prototype acting at a real intersection. The behaviors of two policies are shown: before learning (red dotted line) and after learning (green solid line), in speed, time, and visibility profiles.}
    \label{fig:exp-real-av}
\end{figure}

\subsection{Policy Learning with Real AV}

Our policy learning approach is implemented on a fully operational AV prototype acting in the real world.
We began with uniform policy parameters and drove through an intersection.
At each trip, the action taken by the autonomous agent, and the low-level representation of the observations received by the AV's sensors are recorded at a high frequency.
These data are then used to create policies through \cref{eq:m_sas,eq:m_sas2} for the corresponding intersections.
The intersection has occlusions of parked cars to the left and right, requiring careful edge actions based on the belief.
The AV's speed and time were recorded as it progressed through the intersection.
Then, we drove the AV through a loop of 3 different intersections 5 times.
These experiences were transferred to the scenario used at the original intersection.
This learned policy was then tested at the original intersection.
\cref{fig:exp-real-av} shows the results from these experiments, demonstrating the real-world efficacy of this learning in a deployed autonomous agent.

In \cref{fig:exp-real-av} (b), the speed after the stop line is much slower than the learned policy's speed.
In fact, the initial policy oscillates between stop and edge.
Conversely, the learned policy has learned to cautiously select the edge action until it believes there are no oncoming cars outside of view.
(\cref{fig:exp-real-av} (d) shows the percentage of the side roads that are visible as it traverses.)
The learned policy's improved behavior results in a much smoother and faster navigation, as shown in the time profile in \cref{fig:exp-real-av} (c).

\subsection{Testing in Simulations}
\begin{figure}[!ht]
    \centering
    \subfigure[At intersection $(t=t_0)$]{\includegraphics[width=0.48\columnwidth]{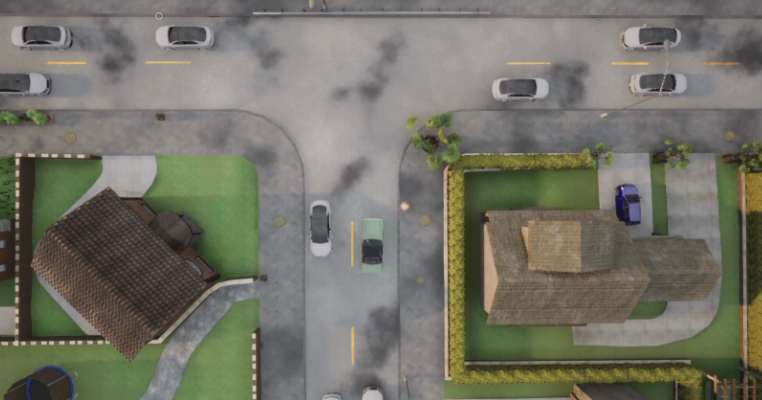}}
    \subfigure[Inside intersection $(t=t_1)$]{\includegraphics[width=0.48\columnwidth]{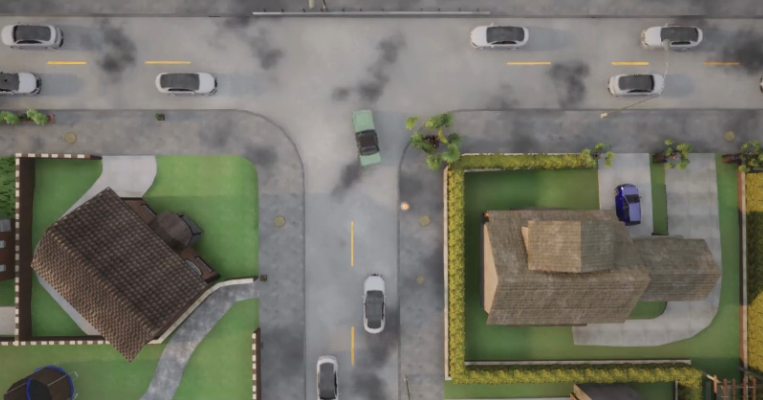}}
    \subfigure[Inside intersection $(t=t_2)$]{\includegraphics[width=0.48\columnwidth]{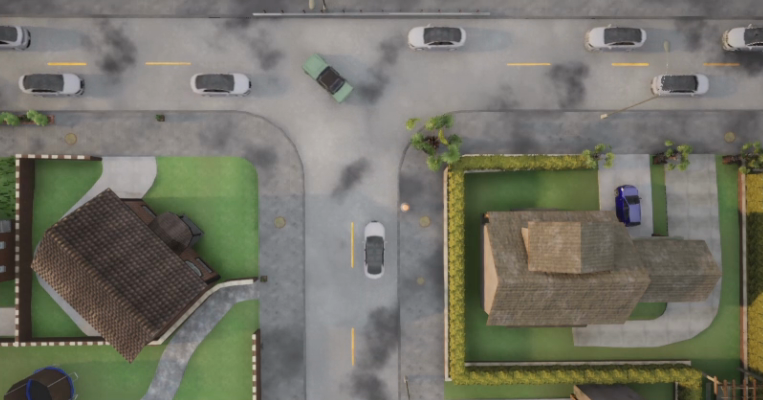}}
    \subfigure[After intersection $(t=t_3)$]{\includegraphics[width=0.48\columnwidth]{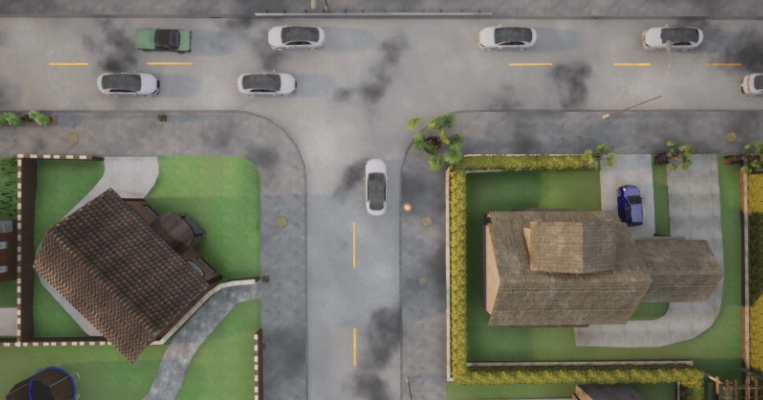}}
    \caption{Ego vehicle (green) navigating around rival vehicles (white) at an unforeseen stop-uncontrolled intersection using the proposed experience filter approach in a CARLA~\cite{carla-sim} simulator environment.}
    \label{fig:intersection-scenario}
\end{figure}

In this setting, we narrow our focus to stop-uncontrolled T-intersections, simulated in the high-fidelity CARLA environment \cite{carla-sim}.
The ego vehicle is controlled by a stop sign before crossing the intersection, and the rival vehicles are uncontrolled, as depicted in \cref{fig:intersection-scenario}.

\subsubsection{Scenario Setup}
We use the following three parameters to represent an environment, which effectively describe the overall characteristics of an intersection:
\begin{align*}
    \psi_1 & \triangleq \text{Corner visibility} \\
    \psi_2 & \triangleq \text{Traffic density} \\
    \psi_3 & \triangleq \text{Driver behavior}
\end{align*}
The corner visibility parameter $ \psi_1 \in \{\texttt{yes, no}\} $ describes whether environmental occlusions (e.g., buildings or parked cars occluding the view of the road) exists around the corner or not.
Traffic density parameter $ \psi_2 \in \{\texttt{low, med, high}\}$ changes depending on the number of the cars in the environment. 
Driver behavior $ \psi_3 \in \{\texttt{cautious, normal, aggressive}\}$  depends on the overall rival speed profiles, for example, an intersection near a school district will likely have vehicles approach the intersection cautiously, whereas a junction to a highway will have aggressive (speeding) vehicles passing through.
In this study, we assume that the environment/task state-space (i.e. $\Space{X}$) is fully observable. This is a reasonable assumption since these characteristics are either static, or can be determined with high confidence. E.g. the congestion of data received through the AV's LIDAR can determine the traffic density.
However, what is unobservable is \textit{how} the other vehicles will react to the ego vehicle, and this is described by the state-space of the POMDP instantiations for all other vehicles in the vicinity.

\begin{table*}[t!]
\centering
\caption{Our experience filter (EF) approach is tested for multiple training efforts on three different metrics: collision risk, discomfort, and time taken (lower is better for all). 
As the training effort increases, the performance of the EF converges to the explicitly trained policy, and outperforms both benchmarks.}
\label{tab:results}
\begin{tabular}{@{} l c c c r @{}} 

    \toprule
    $ $ &  \makecell{Training Effort} & \makecell{Collision Risk} & Discomfort & \makecell{Time Taken} \\
    \midrule
    \textit{Our method} \\
    Experience Filter 
         & $3$ & $0.8173\pm0.1266 $ & $\boldsymbol{0.7483\pm0.0531}$ & $\boldsymbol{0.7971\pm0.0444}$\\
    $ $  & $6$ & $0.5640\pm0.1581$ & $0.5936\pm0.0411$ & $0.6783\pm0.0179$ \\
    $ $  & $9$ & $0.5609\pm0.1554$ & $\boldsymbol{0.4283\pm0.0202}$ & $\boldsymbol{0.5042\pm0.0153}$ \\
    $ $  & $12$ & $\boldsymbol{0.5132\pm0.1369}$ & $\boldsymbol{0.4305\pm0.0206}$ & $\boldsymbol{0.5137\pm0.0125}$ \\
    $ $  & $15$ & $\boldsymbol{0.4587\pm0.1473}$ & $\boldsymbol{0.4345\pm0.0202}$& $\boldsymbol{0.5324\pm0.0325}$ \\
    \midrule
    \textit{Benchmarks} \\
    Entire Dataset 
        & $3$ & $0.7998\pm0.1346$ & $0.8684\pm0.0727$ & $0.9380\pm0.0619$ \\
    $ $ & $6$ & $0.7484\pm0.1517$ & $0.9019\pm0.0805$ & $0.9146\pm0.0555$\\
    $ $ & $9$ & $0.8113\pm0.1297$ & $0.9218\pm0.0781$ & $0.9144\pm0.0528$ \\
    $ $ & $12$ & $0.6838\pm0.1660$ & $0.8762\pm0.0753$ & $0.9127\pm0.0539$ \\
    $ $ & $15$ & $0.8488\pm0.1218$ & $0.9098\pm0.0683$ & $0.9120\pm0.0456$ \\
    $ $ \\
    Nearest Neighbor
        & $3$ & $\boldsymbol{0.7497\pm0.1437}$ & $0.8292\pm0.0620$ & $0.8373\pm0.0484$ \\
    $ $ & $6$ & $\boldsymbol{0.4135\pm0.1635}$ & $\boldsymbol{0.5203\pm0.0416}$ & $\boldsymbol{0.6158\pm0.0291}$ \\
    $ $ & $9$ & $\boldsymbol{0.5201\pm0.1601}$ & $0.5065\pm0.0228$ & $0.5937\pm0.0314$ \\
    $ $ & $12$ & $0.5610\pm0.1462$ & $0.5054\pm0.0232$ & $0.6384\pm0.0421$ \\
    $ $ & $15$ & $0.5411\pm0.1547$ & $0.4643\pm0.0222$ & $0.5643\pm0.0228$ \\
    \midrule
    \textit{Lower Bound} \\
    Explicit Training  
    $ $  & $-$ & $0.4345\pm0.1644$ & $0.4343\pm0.0153$ & $0.5061\pm0.0124$ \\
    \bottomrule

\end{tabular}
\end{table*}

\subsubsection{Policy Learning from Training Data}
\begin{figure}[!ht]
    \centering
    \hspace{1.7cm}\includegraphics[width=0.4\columnwidth]{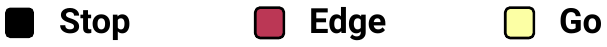}
    \vspace{-0.5px}
    \newline
    \subfigure
    [Policy learnt for T-intersection having high visibility, medium traffic density, normal driver behavior. \label{fig:policy-graph-high-obs}]
    {\includegraphics[width=0.48\columnwidth]{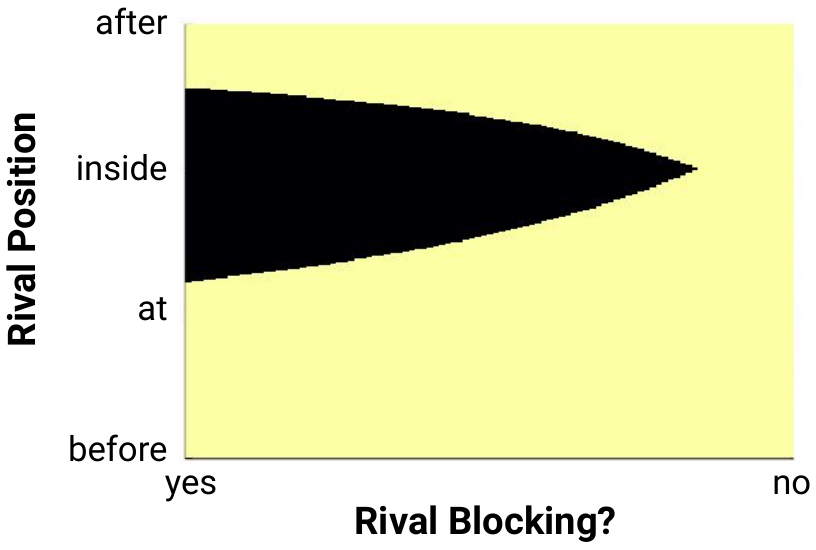}}
    \hspace{2pt}
    \subfigure
    [Policy learnt for T-intersection having low visibility, high traffic density, aggressive driver behavior. \label{fig:policy-graph-low-obs}]
    {\includegraphics[width=0.48\columnwidth]{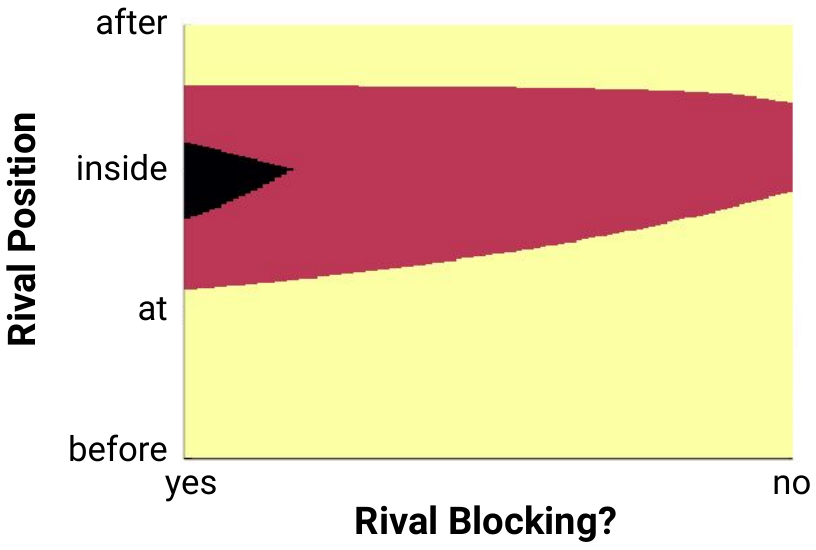}}
    \caption{Example policies learnt from the data collected inside the CARLA simulator, and used with the experience filter created. Ego vehicle is at the stop sign for both policies. As an example, a point in the centroid of the plot would correspond to the belief of the rival car blocking with $50\%$, and being located either ``inside'' or ``at'' the intersection with equal probability.}
    \label{fig:policy-graphs}
    \vspace{-10pt}
\end{figure}

During the design time, the ego vehicle is tasked to make a turn at T-intersections with varying $\boldsymbol\psi = \{\psi_1,\psi_2,\psi_3\}$ parameters that describe the intersection.
The environment state space is consists of every combination of $\psi_1,\psi_2,\psi_3$, hence, there are 18 possible T-intersection characteristics to train on.
Each environment state is run for $101$ scenarios with where the start and goal locations of rival agents are initialized randomly.
Then, similar to the previous section, actions and observations of the AV are recorded, and the policy for each T-intersection characteristic is computed by solving \cref{eqn:pi_star,eqn:V_ba,eqn:Q_sa} using QMDP~\cite{hauskrecht2000value}, where two sample policy maps are given in \cref{fig:policy-graphs}. 


\subsubsection{Creating the Experience Filter}
In this study, we have selected the experience kernel to be a normalized Gaussian kernel:
\begin{align}
    \zeta(x_n, x_k; \sigma, \ell) = \eta \
    \sigma^2
    \exp\left(
    -\dfrac{\Vert x_n-x_k \Vert^2}{2\ell^2}
    \right)
\end{align}
where $\eta$, $\sigma$, and $\ell$ constants are the normalizing factor, kernel variance, and kernel lengthscale, respectively.
These constant values may be identified through likelihood maximization with respect to the data.
Domain knowledge may also be incorporated to the selection of the experience kernel, if available.


\subsubsection{Benchmarking}
We compare our experience filter approach with these three baselines:
\begin{itemize}
    \item {\textit{Entire dataset}: The entire training dataset collected so far is used to learn a a policy, to be benchmarked on the test scenario (i.e., no kernel is used to filter out data).}
    \item {\textit{Nearest neighbor}: The policy used on the test scenario ($\tilde \pi$) whose characteristics are $\tilde \psi$ are through by the policies $\pi_i^*$ trained on $\psi_i$ using the relation:}
\begin{align}
    par(\tilde \pi) & \approxeq par(\pi^*_i) \quad \text{ s.t. } i =  \argmin_{i \in \{ 1, \dots, N\}} \Vert \boldsymbol{\tilde \psi} - \boldsymbol{\psi}_i \Vert
\end{align}
    \item {\textit{Explicit Training}: The transition function for the test scenario is learned explicitly using data collected for this scenario setting. Therefore, this benchmark acts as a lower bound for our performance metrics.}
\end{itemize}
We compare these methods in terms of collision risk, discomfort, and time taken.
The results are shown in \cref{tab:results}.
Here, training effort refers to the number of policies obtained specifically for different training environments, i.e. possible combinations of $\psi_1, \psi_2, \psi_3$.
\textit{Collision risk} is inverse of the minimum distance of the ego vehicle to any rival while navigating through the intersection.
\textit{Discomfort} is integral of the ego vehicle's absolute acceleration during its trajectory.
\textit{Time Taken} is the amount of time taken for the ego vehicle to complete crossing the intersection after stopping at the stop sign.
All three scores are normalized based on the largest value observed over all the testing trials.
As can be seen, our experience filter (EF) approach outperforms both 
the policy trained using the entire dataset
and the nearest neighbor approach,
after a certain amount of training effort. 
The reason we see this trend is due to the fact that as the training effort increases, there is better coverage of the environment state-space $\Space{X}$.
Our results demonstrate that when an unseen task is faced, it is insufficient to naively choose the learned policy of the ``closest-looking" past experience (i.e. \textit{nearest neighbor}). The EF is able to outperform by taking advantage of other experiences through the kernel.
Another critical conclusion is that attempting to use the entire data recorded to create a policy (i.e. \textit{entire dataset}) performs worse than EF.
Intuitively, this is because trying to learn all data at once gives an ``average” performance across different task states $x_i \in \Space{X}$, whereas the EF allows a more intelligent method to correlate 
unseen tasks to past experiences and filter out ones that are less relevant.
We also observe that as the training effort increases, the performance of the experience filter converges to the \textit{explicitly trained} policy, as desired.

\section{Conclusions and Future Work} \label{sec:conclusion}


In this paper, we have presented a novel approach to generate policies for tasks or environments an autonomous vehicle agent has not seen before.
We propose an \textit{experience filter} that utilizes a kernel over parameterized learned policies that had been trained on different past tasks/environments.
This kernel suggests a correlation factor between the learned policy for the unseen task and past experiences, eliminating the requirement of collecting any new data, and thereby allowing fast, low-effort, and near-optimal solutions.

Our methodology assumes both policies and environments can be represented by low-dimensional, easily accessible parameter vectors. 
Tests on both an actual AV and a realistic simulation environment are performed, where the interactions between the ego and other vehicles are modeled as multiple partially observable Markov decision processes (POMDPs). 
Our results indicate that the experience filter yields a higher performance than simply using the entire data collected as commonly done, and is able to filter out irrelevant past experiences.
We also show that our kernel-based approach outperforms a naive nearest neighbor approach.

Future work includes testing the experience filter on other real-life domains beyond T-intersections, and extending our implementation that accommodates continuous dynamics.
We will also look into the optimization
of the kernel itself with respect to existing data, and
inspect under which conditions a kernel would satisfy certain
optimality bounds.

\section*{Code}
The code used for this paper can be found publicly in this GitHub repository: 
\url{https://github.com/sisl/Experience-Filter}

\section*{Acknowledgments}
The research reported in this work was supported by\linebreak
Alliance Innovation Laboratory Silicon Valley, Nissan North America, Inc. We thank Marcell J. Vazquez-Chanlatte for his valuable feedback.


\renewcommand*{\bibfont}{\small}
\printbibliography

@String { aaai        = {AAAI Conference on Artificial Intelligence (AAAI)} }

@String { corl        = {Conference on Robot Learning (CoRL)} }

@String { iclr        = {International Conference on Learning Representations} }

@String { icml        = {International Conference on Machine Learning (ICML)} }

@String { ijcai       = {International Joint Conference on Artificial Intelligence (IJCAI)} }

@String { itsc        = {IEEE International Conference on Intelligent Transportation Systems (ITSC)} }

@String { jair        = {Journal of Artificial Intelligence Research} }

@String { mit         = {Massachusetts Institute of Technology} }

@String { nips        = {Advances in Neural Information Processing Systems (NIPS)} }

@String { rss         = {Robotics: Science and Systems} }

@String {arxiv        = {arXiv} }

@ARTICLE{sunberg_pomdp,
  author={Sunberg, Zachary and Kochenderfer, Mykel J.},
  journal={IEEE Transactions on Intelligent Transportation Systems}, 
  title={Improving Automated Driving Through POMDP Planning With Human Internal States}, 
  year={2022},
  volume={23},
  number={11},
  pages={20073-20083},
  doi={10.1109/TITS.2022.3182687}}

@inproceedings{cailets,
  title={LeTS-Drive: Driving in a Crowd by Learning from Tree Search},
  author={Cai, Panpan and Luo, Yuanfu and Saxena, Aseem and Hsu, David and Lee, Wee Sun},
  booktitle=rss,
  year={2019}
}

@inproceedings{carla-sim,
  title = { {CARLA}: {An} Open Urban Driving Simulator},
  author = {Alexey Dosovitskiy and German Ros and Felipe Codevilla and Antonio Lopez and Vladlen Koltun},
  booktitle = corl,
  pages = {1--16},
  year = {2017}
}

@article{hauskrecht2000value,
  title={Value-function approximations for partially observable {M}arkov decision processes},
  author={Hauskrecht, Milos},
  journal=jair,
  volume={13},
  pages={33--94},
  year={2000}
}

@article{kaelbling1998planning,
  title={Planning and acting in partially observable stochastic domains},
  author={Kaelbling, Leslie Pack and Littman, Michael L and Cassandra, Anthony R},
  journal={Artificial Intelligence},
  volume={101},
  number={1-2},
  pages={99--134},
  year={1998},
}

@book{kochenderfer2015_dmu,
  title={Decision Making Under Uncertainty: Theory and Application},
  author={Kochenderfer, M.J.},
  year={2015},
  publisher={MIT Press}
}

@book{bellman2015applied,
  title={Applied Dynamic Programming},
  author={Bellman, Richard E and Dreyfus, Stuart E},
  year={2015},
  publisher={Princeton University Press}
}

@book{kotz2004continuous,
  title={Continuous Multivariate Distributions},
  author={Kotz, Samuel and Balakrishnan, Narayanaswamy and Johnson, Norman L},
  year={2004},
  publisher={Wiley}
}

@article{zhuang2020comprehensive,
  title={A comprehensive survey on transfer learning},
  author={Zhuang, Fuzhen and Qi, Zhiyuan and Duan, Keyu and Xi, Dongbo and Zhu, Yongchun and Zhu, Hengshu and Xiong, Hui and He, Qing},
  journal={Proceedings of the IEEE},
  volume={109},
  number={1},
  pages={43--76},
  year={2020},
}

@ARTICLE{rusu2016progressive,
       author = {{Rusu}, Andrei A. and {Rabinowitz}, Neil C. and {Desjardins}, Guillaume and {Soyer}, Hubert and {Kirkpatrick}, James and {Kavukcuoglu}, Koray and {Pascanu}, Razvan and {Hadsell}, Raia},
        title = "{Progressive Neural Networks}",
         year = 2016,
archivePrefix = {arXiv},
       eprint = {1606.04671},
 primaryClass = {cs.LG},
       adsurl = {https://ui.adsabs.harvard.edu/abs/2016arXiv160604671R}
}

@inproceedings{rajendran2017attend,
  author    = {Janarthanan Rajendran and
               Aravind S. Lakshminarayanan and
               Mitesh M. Khapra and
               P. Prasanna and
               Balaraman Ravindran},
  title     = {Attend, Adapt and Transfer: Attentive Deep Architecture for Adaptive Transfer from multiple sources in the same domain},
  booktitle = iclr,
  year      = {2017},
}

@inproceedings{taylor2008transferring,
  title={Transferring instances for model-based reinforcement learning},
  author={Taylor, Matthew E and Jong, Nicholas K and Stone, Peter},
  booktitle={European Conference on Machine Learning and Knowledge Discovery in Databases},
  pages={488--505},
  year={2008}
}

@inproceedings{chrisman1992reinforcement,
  title={Reinforcement learning with perceptual aliasing: {T}he perceptual distinctions approach},
  author={Chrisman, Lonnie},
  booktitle=aaai,
  volume={1992},
  pages={183--188},
  year={1992},
}

@article{choi1999environment,
  title={An environment model for nonstationary reinforcement learning},
  author={Choi, Samuel and Yeung, Dit-Yan and Zhang, Nevin},
  journal=nips,
  volume={12},
  year={1999}
}

@article{killian2017robust,
  title={Robust and efficient transfer learning with hidden parameter {M}arkov decision processes},
  author={Killian, Taylor W and Daulton, Samuel and Konidaris, George and Doshi-Velez, Finale},
  journal=nips,
  volume={30},
  year={2017}
}

@inproceedings{perez2020generalized,
  title={Generalized hidden parameter {MDP}s: {T}ransferable model-based {RL} in a handful of trials},
  author={Perez, Christian and Such, Felipe Petroski and Karaletsos, Theofanis},
  booktitle=aaai,
  volume={34},
  number={04},
  pages={5403--5411},
  year={2020}
}

@article{wang2019survey,
  title={A survey of zero-shot learning: Settings, methods, and applications},
  author={Wang, Wei and Zheng, Vincent W and Yu, Han and Miao, Chunyan},
  journal={ACM Transactions on Intelligent Systems and Technology (TIST)},
  volume={10},
  number={2},
  pages={1--37},
  year={2019},
  publisher={ACM New York, NY, USA}
}

@inproceedings{higgins2017darla,
  title={Darla: Improving zero-shot transfer in reinforcement learning},
  author={Higgins, Irina and Pal, Arka and Rusu, Andrei and Matthey, Loic and Burgess, Christopher and Pritzel, Alexander and Botvinick, Matthew and Blundell, Charles and Lerchner, Alexander},
  booktitle=icml,
  pages={1480--1490},
  year={2017},
}

@inproceedings{ammar2014automated,
  title={An automated measure of {MDP} similarity for transfer in reinforcement learning},
  author={Ammar, Haitham Bou and Eaton, Eric and Taylor, Matthew E and Mocanu, Decebal Constantin and Driessens, Kurt and Weiss, Gerhard and Tuyls, Karl},
  booktitle={Workshops at the AAAI Conference on Artificial Intelligence},
  year={2014}
}

@ARTICLE{shu-itvt-2022,
  author={Shu, Hong and Liu, Teng and Mu, Xingyu and Cao, Dongpu},
  journal={IEEE Transactions on Vehicular Technology}, 
  title={Driving Tasks Transfer Using Deep Reinforcement Learning for Decision-Making of Autonomous Vehicles in Unsignalized Intersection}, 
  year={2022},
  volume={71},
  number={1},
  pages={41-52},
  doi={10.1109/TVT.2021.3121985}}

@inproceedings{xu2018zero,
  title={Zero-shot deep reinforcement learning driving policy transfer for autonomous vehicles based on robust control},
  author={Xu, Zhuo and Tang, Chen and Tomizuka, Masayoshi},
  booktitle=itsc,
  pages={2865--2871},
  year={2018},
}

@incollection{harrison2020adapt,
  title={Adapt: zero-shot adaptive policy transfer for stochastic dynamical systems},
  author={Harrison, James and Garg, Animesh and Ivanovic, Boris and Zhu, Yuke and Savarese, Silvio and Fei-Fei, Li and Pavone, Marco},
  booktitle={Robotics Research},
  pages={437--453},
  year={2020},
  publisher={Springer}
}

@inproceedings{ijcai2017-664,
  author    = {Kyle Hollins Wray and Stefan J. Witwicki and Shlomo Zilberstein},
  title     = {Online Decision-Making for Scalable Autonomous Systems},
  booktitle = ijcai,
  pages     = {4768--4774},
  year      = {2017},
  doi       = {10.24963/ijcai.2017/664},
  url       = {https://doi.org/10.24963/ijcai.2017/664},
}
\end{document}